\documentclass[letterpaper, 10 pt, conference]{ieeeconf}  % Comment this line out if you need a4paper

\IEEEoverridecommandlockouts                              % This command is only needed if 
\usepackage{booktabs}
\usepackage{times}
\usepackage{epsfig}
\usepackage{cite}
\usepackage{graphicx}
\usepackage{amsmath}
\usepackage{amssymb}
\usepackage[pagebackref,breaklinks,colorlinks]{hyperref}

\title{\LARGE \bf
ReCoAt: A Deep Learning-based Framework for Multi-Modal Motion Prediction in Autonomous Driving Application
}

\author{Zhiyu Huang, Xiaoyu Mo, and Chen Lv$^{*}$,~\IEEEmembership{Senior Member, IEEE} %
\thanks{Zhiyu Huang, Xiaoyu Mo, and Chen Lv are with the School of Mechanical and Aerospace Engineering, Nanyang Technological University, 639798, Singapore. (E-mails: \{zhiyu001, xiaoyu006\}@e.ntu.edu.sg, lyuchen@ntu.edu.sg)}%
\thanks{This work was supported in part by A*STAR Grant (No. W1925d0046), A*STAR AME Young Individual Research Grant (No. A2084c0156), and SUG-NAP Grant of Nanyang Technological University, Singapore.}%
\thanks{$^{*}$Corresponding author: C. Lv}%
}

\begin{document}

\maketitle
\thispagestyle{empty}
\pagestyle{empty}

%%%%%%%%%%%%%%%%%%%%%%%%%%%%%%%%%%%%%%%%%%%%%%%%%%%%%%%%%%%%%%%%%%%%%%%%%%%%%%%%
\begin{abstract}
This paper proposes a novel deep learning framework for multi-modal motion prediction. The framework consists of three parts: recurrent neural network to process target agent's motion process, convolutional neural network to process the rasterized environment representation, and distance-based attention mechanism to process the interactions among different agents. We validate the proposed framework on a large-scale real-world driving dataset, Waymo open motion dataset, and compare its performance against other methods on the standard testing benchmark. The qualitative results manifest that the predicted trajectories given by our model are accurate, diverse, and in accordance with the road structure. The quantitative results on the standard benchmark reveal that our model outperforms other baseline methods in terms of prediction accuracy and other evaluation metrics. The proposed framework is the second-place winner of the 2021 Waymo open dataset motion prediction challenge.

\end{abstract}

%%%%%%%%%%%%%%%%%%%%%%%%%%%%%%%%%%%%%%%%%%%%%%%%%%%%%%%%%%%%%%%%%%%%%%%%%%%%%%%%
\section{Introduction}
Decision-making is one of the most crucial and challenging tasks for autonomous driving to operate safely among human drivers without causing inconvenience and dangers \cite{cyber_attack, av_survey, future_iv, cognitive_design,yang_jas}. Making human-like and intelligent decisions has gained a lot of research interest recently because driving is a highly interactive task that involves proactively reasoning about other agents’ actions and intentions, and how to make informed decisions still remains an open and challenging problem for autonomous \cite{hao_tvt, hu_vtm, ais, hu_tiv, he_tiv,peng_tmech}. Solving this problem will trigger the next major wave of success for driving automation and substantially propel the application of autonomous vehicles (AVs). Therefore, many works employ the learning-based approaches to tackle the decision-making problems because they are scalable and able to handle complex real-world scenarios. Such methods include the likes of deep imitation learning \cite{huang2020multi}, deep reinforcement learning \cite{huang2022efficient, wu2021human, wu2021uncertainty, wu2021prioritized, liu2021improved} and inverse reinforcement learning \cite{huang2021driving}. However, the main drawback of the black-box driving policies trained from such learning-based methods is the lack of interpretability, reliability, and robustness.

On the other hand, a promising way is through motion prediction and planning \cite{huang2021multi, mo2022multi}. By predicting the intentions or future states of the surrounding traffic participants, we can make decisions accordingly in an interpretable, reliable, and intelligent way. Therefore, accurately predicting traffic participants' future trajectories is an essential part of this framework, especially in crowded and complex traffic scenarios. However, it is a remarkably challenging task due to the following factors. First and foremost, in addition to the dynamics or physical constraints, the future trajectories of the traffic participants are highly dependent on the environment information. This encompasses the information of the map (e.g., drivable lanes for vehicles and crosswalks for pedestrians), traffic signals (e.g., traffic lights and stop signs), traffic rules, and more importantly the interactions among different traffic participants. Another influential factor is the uncertainties in intention and behavior. There may be a variety of plausible future motion trajectories for the target due to unknown destinations and noisy movements. This requires the motion prediction model to be able to output multiple possible trajectories and their likelihoods. 

Facing the two major challenges, we propose a deep learning framework that can output multi-modal predictive trajectories by modeling the interaction between different agents and exploiting the environmental context information, as well as the dynamics information of the target agent. Specifically, we utilize a convolutional neural network (CNN) to encode environmental context information, which is rasterized as a bird-eye-view image to represent the road structure, traffic signals, and other agents' positions. A recurrent neural network (RNN) is used to encode the historical states of the interacting agents and a novel attention mechanism is proposed to fuse the information of interacting agents to represent their interactions with the target agent. The main contributions of this paper is summarized as follows:
\begin{itemize}
    \item A deep neural network framework is proposed, consisting of Recurrent, Convolution, and Attention (ReCoAt) operations, to predict the multi-modal futures of the target agent based on the agent's historical states, environment context, and interactions.
    \item The proposed framework is validated on a large-scale real-world urban driving dataset and shows competitive prediction accuracy compared to baseline methods on the standard testing benchmark\footnote{https://waymo.com/open/challenges/2021/motion-prediction/}.
\end{itemize}

\section{Related work}
Motion prediction is a long-researched research area. One of the simplest (but sometimes effective) approaches is physics-based models, such as the constant velocity model, constant acceleration model, and constant yaw rate and velocity model. However, dynamics models cannot account for the road environment or interactions among agents, and thus some approaches focus on mathematically formulating the interactions between, such as the intelligent driver model \cite{treiber2000congested} and social forces model \cite{luber2010people}. These model-based methods are computationally-efficient and easy to implement, but their performance and accuracy are limited. More recently, learning-based methods have gained great interest and a large body of literature has applied deep neural networks to motion prediction \cite{cui2019multimodal, zhao2020tnt, huang2021multi, mo2022multi}. They rely on a wealth of observation data to capture the complexity of road structures and interactions among multiple agents encountered in real-world environments, exhibiting excellent prediction accuracy and generalization ability.

At first, researchers usually formulate the motion prediction as a time series prediction problem, which is to use a sequence of historical states to predict a sequence of future states. Therefore, RNN, or more particularly, long short-term memory (LSTM) networks have been widely applied to motion prediction \cite{Alahi_2016_CVPR}. To incorporate the environment information (e.g., road structure, traffic signals) into motion prediction, many works employ the image structure that rasterizes the driving environment into 2D grids with each pixel representing a semantic class, which can be effectively processed by CNNs \cite{cui2019multimodal, gilles2021home, chai2019multipath}. However, the interaction between traffic participants is not explicitly represented in the rasterized images. As for explicitly modeling the interactions, many recent works attempt to use graph modeling and the attention mechanism, because the attention-based feature fusion can essentially represent and model the interaction between agents. For example, \cite{mercat2020multi} utilize multi-head self-attention to account for interactions among different vehicles, and \cite{dong2021multi} and \cite{gao2020vectornet} propose to utilize graph attention networks to extract relational features on the scene graph containing different agents. Considering that in a traffic scenario, the attention the target agent pays to a surrounding agent is largely determined by the distance between them, we propose a distance attention module to model the interaction between the target agent and its surrounding agents.

While the deep learning-based methods enjoy strong performance, a subtle point is still essential for safety-critical applications such as autonomous driving, which is to predict multiple possible future trajectories, ideally with the likelihoods of each occurring, in order to make safe decisions. To address this issue, some works propose to use generative models, such as conditional variational autoencoder \cite{salzmann2020trajectron++}. However, such methods may require thousands of samples to recover a meaningful distribution, which could significantly slow down the inference speed. An alternative approach is to model the multimodal distribution over future trajectories as a Gaussian mixture model (GMM) and the network is used to predict the parameters of the GMM. However, using the ground truth as one certain trajectory while predicting diverse output trajectories suffers from mode collapse problem \cite{rhinehart2018r2p2}, which means the distribution collapses to one single mode. To address this problem, we set up an ensemble of trajectory decoders and follow the training method in \cite{cui2019multimodal} by only training the decoder of the closest mode to the ground-truth trajectory, which could ensure the diversity of the predicted trajectories and stabilize training. We also add a classification branch in the network to predict a confidence score or a normalized probability for each predicted trajectory.

\section{The High-Level Framework}
\subsection{Problem formulation}
The task of motion prediction is to predict the possible future trajectories of a target agent over a time horizon $T_f$ based on its historical states over a time period $T_h$ and environmental context information. The input $\mathbf{X}$ to the prediction model consists of the historical dynamic states of itself ($S_{0}$) and its surrounding agents ($S_{1}, \dots, S_{N}$), as well as the current environment information $\mathcal{M}$. Without loss of generality, we assume that there are $N$ surrounding agents around the target agent, however, the number of surrounding agents can be varied in different situations. The agents include vehicles, pedestrians, and cyclists. The output of the prediction model $\Hat{\mathbf{Y}}$ is $K$ trajectories, each consisting of a sequence of 2D coordinates denoting the possible future positions of the target agent. Mathematically, the problem is formulated as:
\begin{equation}
\begin{aligned}
\mathbf{X} &= \{ S_{0}, S_{1}, \dots, S_{N}, \mathcal{M} \}, \\
\Hat{\mathbf{Y}} &= \{ (x^t_{j}, y^t_{j}) | t \in \{ t_{0}+1, \dots, t_{0}+T_{f} \} \}_{j=1}^{K},
\end{aligned}
\end{equation}
where $S_{i}=\{ s_{i}^{t_0-T_h+1}, s_{i}^{t_0-T_h+2}, \dots, s_{i}^{t_0} \}$ and $s_{i}^{t}$ is the dynamic state of the agent $i$ at timestep $t$, $(x^t_{j}, y^t_{j})$ is the $j$th predicted coordinate of the target agent at timestep $t$, and $t_0$ is the current time step.

\subsection{Dataset and data processing}
We employ the Waymo open motion dataset \cite{ettinger2021large}, which is a large-scale and diverse motion forecasting dataset that contains over 100,000 driving scenes with interesting interactions between vehicles, pedestrians, and cyclists. The dataset gives agents' tracks for the past 1 second at a 10Hz sampling rate and a corresponding map, and the motion prediction task is to predict the future positions of target agents for 8 seconds with a sampling rate of 2Hz. 

The historical dynamic state of the target agent and its surrounding agents $s_{i}^{t}$ is in the format of $(x, y, v_x, v_y, \theta)$, where $(x, y)$ is the coordinate, $(v_x, v_y)$ the velocity, and $\theta$ the heading angle. Note that the coordinate system is centered on the target agent's position at the current timestep with its heading aligned with the x-axis. Thus, for each agent, its state representation is a tensor with shape $(10, 5)$. We only consider up to ten surrounding agents within a radius of 30 meters to the target agents and incorporate the states of them into a fixed-length tensor with shape $(10, 10, 5)$. The surrounding agents in the tensor are ordered according to their distances to the target agent and vacancies in the tensor are padded with zeros if there are not enough surrounding vehicles found.

\begin{figure}[htp]
\centering
\includegraphics[width=0.95\linewidth]{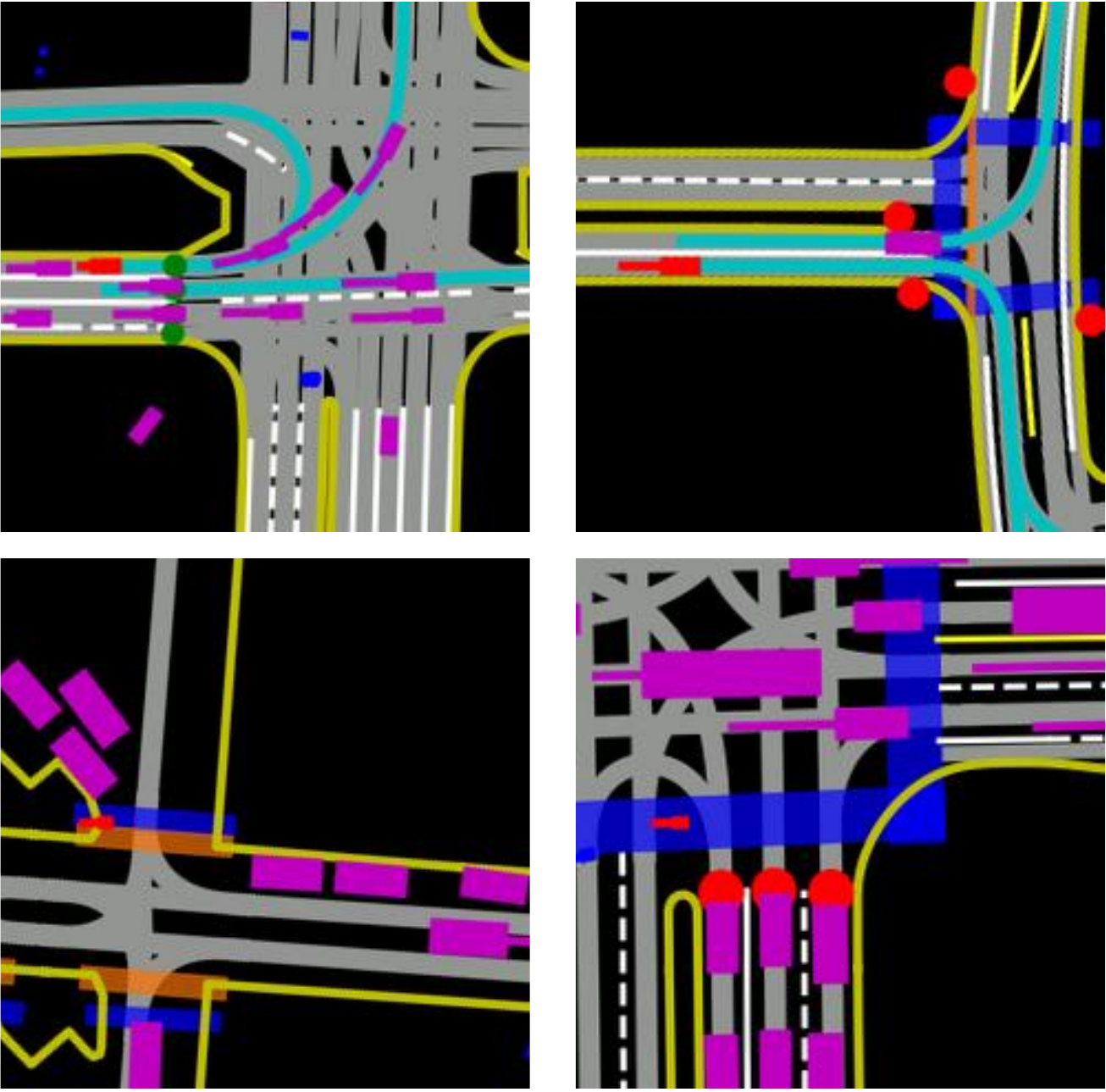}
\caption{Examples of rasterized environment representation}
\label{fig1}
\end{figure}

The environmental information is represented as bird-eye-view rasterized 2D images, as shown in Fig. \ref{fig1}. The target agent represented as a red box is positioned at the position $(1/5, 1/2)$ of the image. The surrounding agents are shown in different colors: magenta for vehicles, blue for pedestrians, and green for cyclists. The tails (thin lines attached to the boxes) behind the agents are the historical tracks. The drivable lanes are displayed as grey polygons and the candidate centerlines for the target agent are painted in cyan. The candidate centerlines in the format of polylines (sequence of waypoints) are also used as additional information for predicting the vehicle's future motion. The red circles or green circles on the road show the states of the traffic lights and the red circles on the roadside represent the stop signs. The blue polygon is the pedestrian crossing and the orange polygon is the speed bump. The lane markings are represented in different kinds of polylines: yellow solid lines for road edges, white solid lines for solid white road lines, white dashed lines for broken white road lines, light yellow lines for yellow road lines. The bird-eye-view rasterized images are in the size of $240\times240\times3$ encoding different sizes of areas for different types of target agents. For vehicles, the actual area in the scene is $80m \times 80m$, and $60m \times 60m$ for cyclists and $40m \times 40m$ for pedestrians. 

\subsection{Model: ReCoAt}

\begin{figure*}[htp]
\centering
\includegraphics[width=0.95\linewidth]{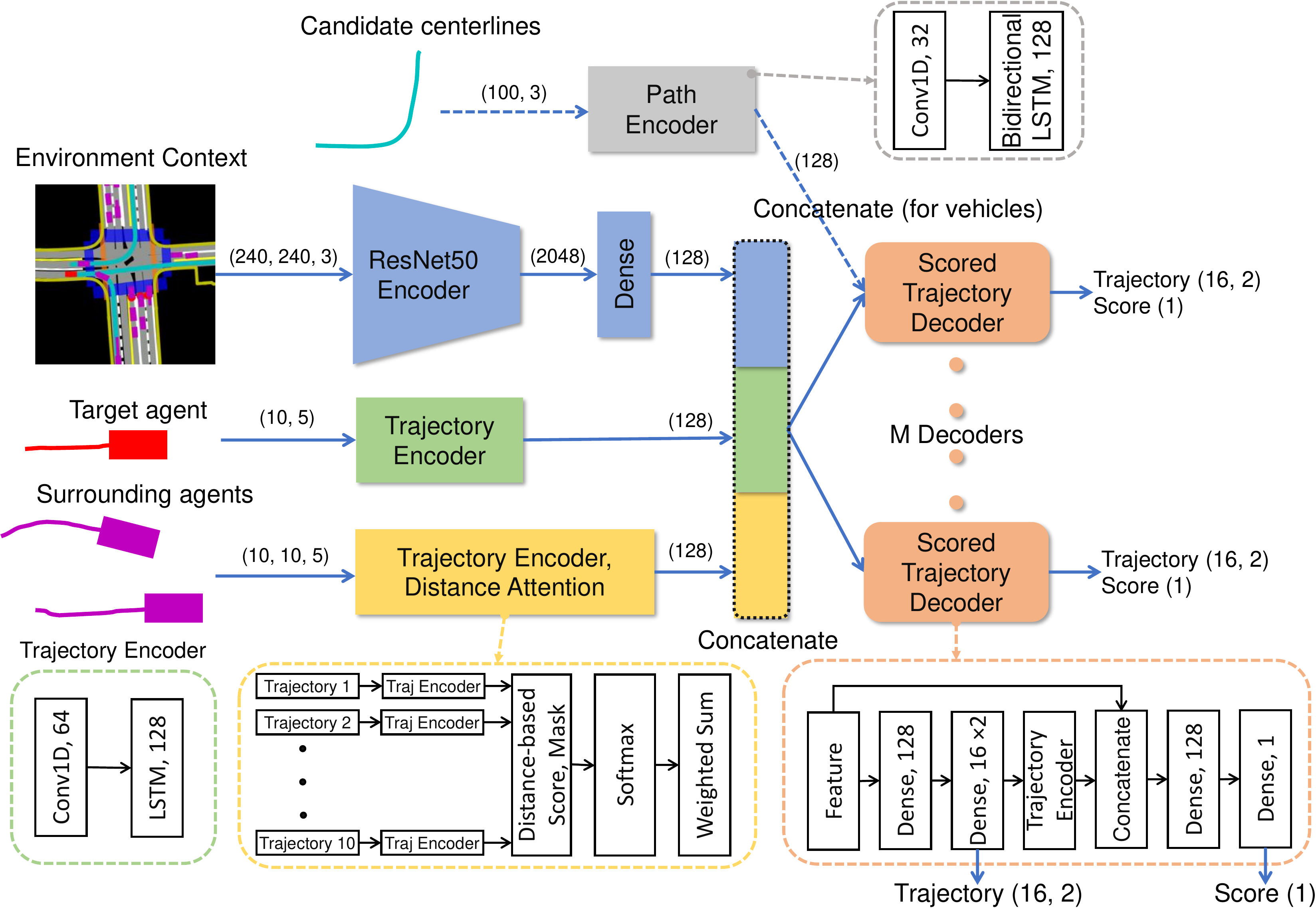}
\caption{The structure of the ReCoAt motion prediction model}
\label{fig2}
\end{figure*}

The structure of the motion prediction model named ReCoAt is visualized in Fig. \ref{fig2}. The name comes from different parts of the model, which are recurrent neural network to process the target agent's dynamic state, convolutional neural network to process environmental context, and attention module to process the interactions among agents. The environment representation in the format of bird-eye-view rasterized images is processed by a ResNet50 encoder \cite{he2016identity}, which outputs a 2048-dimension feature vector, followed by a fully connected layer to reduce its dimension to 128. The past trajectory of the target agent is processed by a trajectory encoder, which consists of a 1D convolutional layer and an LSTM layer, to extract the motion information of the target agent. The output of the trajectory encoder is a 128-dimension feature vector. To model the interaction between the target agent and its surrounding agents, the proposed distance attention module is employed and illustrated below.

First of all, all the surrounding agents' trajectories are projected to high-dimensional feature vectors by trajectory encoders with shared weights across them. In accordance with the definition of the attention mechanism, the query is the current position of the target agent, keys are the positions of the surrounding agents, values are the feature vectors obtained by the trajectory encoder. The score function, which rates which agents the target agent needs to pay attention to, is defined as:
\begin{equation}
f_{att} (\text{key}_i, \text{query}) = \frac{\alpha}{\sqrt{(x_i-x_0)^2+(y_i-y_0)^2}},
\end{equation}
where $\alpha$ is a hyper parameter, $(x_0, y_0)$ is the position of the target agent, and $(x_i, y_i)$ is the position of the surrounding agent $i$. The intuition behind this score is that the target agent needs to pay more attention to agents that are closer to it and less attention to those that are still far away. 
The attention weights are calculated by a softmax over all score function outputs:
\begin{equation}
\beta_i = \frac{\exp f_{att} (\text{key}_i, \text{query}) }{\sum_j \exp \left( f_{att} (\text{key}_j, \text{query}) \right) }.
\end{equation}
The output of the attention module is calculated as the weighted sum of the value vectors:
\begin{equation}
\text{out} = \sum_{i} \beta_{i} \cdot \text{value}_{i}.
\end{equation}
In practice, we need to mask out the padding elements by assigning the corresponding score function outputs with a large negative value.

Concatenating the feature vectors from the target agent, environment context, and agent interaction, we obtain a low-dimensional feature vector. For vehicles, we also add the information of candidate centerlines processed by a path encoder to the feature vector. We use a mixture of different trajectory decoders to output multi-modal trajectories and their associated scores. The trajectory is given by two dense layers representing $x$ and $y$ coordinates respectively. The scoring of the trajectory is conditioned on the feature vector and the encoded trajectory feature. The scores of all the predicted trajectories are then stacked and passed through a softmax layer to output the confidence score (probabilities) of these trajectories. The diversity among the trajectory decoders is encouraged by only updating the trajectory regression part of the winning decoder, of which the output trajectory is the closest to the ground truth, while the scoring part for all the decoders can be updated. This could help assign each training example to a particular mixture and also each mode to specialize for a distinct class of behaviors (e.g., going straight and turning) \cite{cui2019multimodal}.

\subsection{Loss function}
To improve the model's performance in long-term prediction and low-speed prediction, we use the weighted mean absolute error whose weights scale with time $t$ and the initial speed of the target agent $v$. For a data point, its trajectory loss is defined as:
\begin{equation}
\mathcal{L}_{traj} = \min_{j \in \{1, \dots, K\}} \frac{1}{T_f} \sum_t w_{t}^{v} \parallel (x_{gt}^{t}, y_{gt}^t) - (x_{j}^{t}, y_{j}^t) \parallel,
\end{equation}
where $(x_{gt}^{t}, y_{gt}^t)$ is the ground truth position at time step $t$. The per-step weight $w_{t}^{v}$ is defined as follows:
\begin{equation}
w_{t}^v = w_t \cdot w^v, \ w_t = 0.5t, \ w^v = \max(1, 4-0.2v).
\end{equation}

The scoring loss is the cross entropy loss between the ground truth probability distribution and the predicted distribution. For a data point, the scoring loss is defined as:
\begin{equation}
\mathcal{L}_{score} = \mathcal{L}_{CE}(p_{gt}, p),
\end{equation}
where $p$ is the predicted probability distribution and the ground truth distribution $p_{gt}$ is defined according to the L2 distance between the trajectory endpoint $\mathbf{s}_{}^{T_f}$ and ground truth endpoint $\mathbf{s}_{gt}^{T_f}$:
\begin{equation}
p_{gt} = \frac{\exp -\parallel \mathbf{s}_{}^{T_f} - \mathbf{s}_{gt}^{T_f} \parallel_2}{\sum_j \exp -\parallel \mathbf{s}_{j}^{T_f} - \mathbf{s}_{gt}^{T_f} \parallel_2}.    
\end{equation}

The total loss is a weighted sum of the trajectory regression loss and the scoring loss:
\begin{equation}
\mathcal{L} = \mathcal{L}_{score} + \lambda \mathcal{L}_{traj},
\end{equation}
where $\lambda$ is a hyperparameter to balance the scales of loss terms in the training process.

\subsection{Implementation details}
The framework is implemented with TensorFlow and trained on two NVIDIA RTX 2080Ti GPUs. All the activation functions in the dense layers are the ELU, and all the dense layers (except the final output layers) are followed by dropout layers (with a dropout rate of 0.5) to mitigate overfitting. The omitted parameters of other layers follow the default settings in Tensorflow. The number of ensemble decoders is set to six, which means the framework can output six possible future trajectories and their scores. The parameter $\alpha$ in the attention module is set to 10 and the parameter $\lambda$ in the loss function is set to 0.2. Different types of target agents (vehicles, cyclists, and pedestrians) are grouped together to train different type-specific prediction models. The total training data for vehicles is 1,891,251, 244,748 for pedestrians, and 86,775 for cyclists. We use Nadam optimizer with a learning rate that starts with $3e^{-4}$ and decays by a factor of 0.9 after every epoch. The number of training epochs is 50 and the batch size is 32.

\section{Results and discussions}
\subsection{Qualitative results}
\begin{figure}[htp]
\centering
\includegraphics[width=\linewidth]{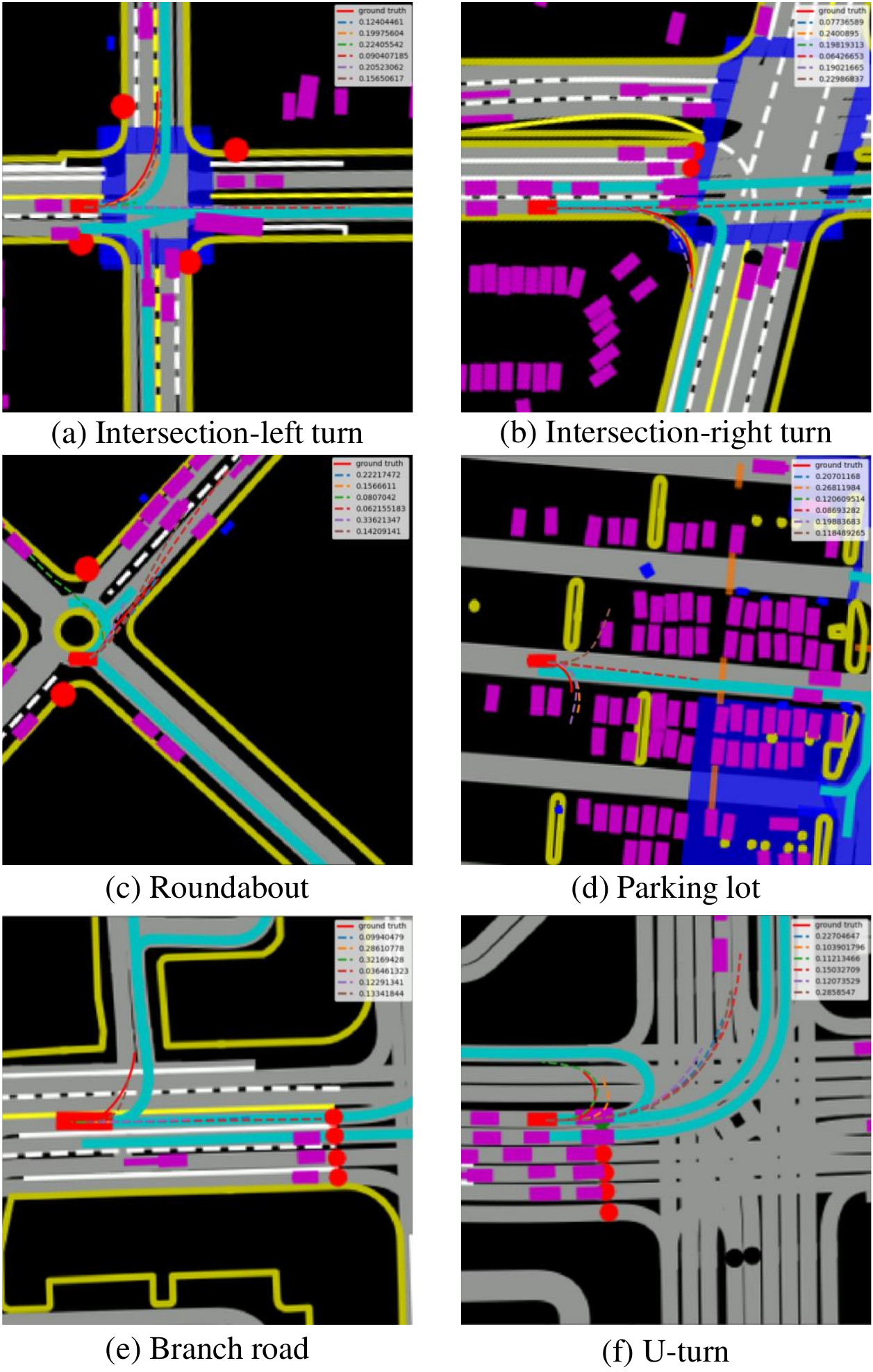}
\caption{Examples of the model predictions}
\label{fig3}
\end{figure}

Fig. \ref{fig3} shows some examples of multi-modal motion prediction given by our framework, covering a wide variety of scenarios. Six possible trajectories along with their probabilities are displayed, as well as the ground truth trajectory. The results reveal that the predicted trajectories are diverse with high coverage of different possible behaviors and in accordance with the road structures, which shows the accuracy, diversity, and map-adaptability of our proposed framework in different traffic scenarios. Moreover, those trajectories that are closer to the ground truth are scored higher given by the framework, which indicates the excellent capability of the framework at evaluating the likelihoods of the predicted trajectories, which is crucial from the downstream planning or decision-making module to produce safe and human-like driving decisions.

\subsection{Quantitative results}
Table \ref{tab1} gives the results of the commonly-used evaluation metrics of our model on the standard test benchmark of the Waymo motion dataset, in compassion with the other baseline methods (with rasterized image scene representations) from the leaderboard. The metrics used are minimum average displacement error (minADE), minimum final displacement error (minFDE), miss rate, overlap rate, and mean average precision (mAP). Detailed definitions of these metrics can be found on the motion prediction challenge website\footnote{https://waymo.com/open/challenges/2021/motion-prediction/}. The results listed in Table \ref{tab1} are averaged over different object types (vehicles, pedestrians, cyclists) and different evaluation times (3, 5, and 8 seconds). mAP is the primary ranking metric as it gauges not only the accuracy of predictions but also the confidence value (scoring) of predictions. The results in Table \ref{tab1} reveal that our model shows the highest mAP value among the baseline methods and comparable performance on other metrics, which means the proposed framework has improved prediction accuracy in both position error and scoring compared to the baseline methods. Also, using only historical state sequences and LSTM encoder performs the worst, which indicates that adding the environment context (road structures, traffic signals, interactions, etc) is beneficial to improve the accuracy and generalization ability.

\begin{table}[htp]
\caption{Comparison of average metrics on the standard testing benchmark of the Waymo motion dataset}
\resizebox{\linewidth}{!}{%
\begin{tabular}{@{}cccccc@{}}
\toprule
\multicolumn{1}{l}{Method} &
  \multicolumn{1}{l}{minADE (m) $\downarrow$} &
  \multicolumn{1}{l}{minFDE (m) $\downarrow$} &
  \multicolumn{1}{l}{Miss Rate $\downarrow$} &
  \multicolumn{1}{l}{Overlap Rate $\downarrow$} &
  \multicolumn{1}{l}{mAP $\uparrow$} \\ \midrule
CNNOnRaster     & \textbf{0.7400} & \textbf{1.4936} & \textbf{0.2091} & 0.1640          & 0.2136                 \\
AIR             & 0.8682          &  1.6691         & 0.2333          & \textbf{0.1583} & 0.2596                \\
MultiPath \cite{pmlr-v100-chai20a}& 0.7430          &  1.6612         & 0.2475          & 0.1584          & 0.2614               \\
Ours            & 0.7703          &  1.6668         & 0.2437          & 0.1642          & \textbf{0.2711} \\ \bottomrule
\end{tabular}
}
\label{tab1}
\end{table}

\subsection{Discussions}
The proposed ReCoAt motion prediction framework has shown outstanding performance by taking into account the agent's dynamics, environment context, and interactions between agents and processing the information with appropriate neural operations. Nonetheless, one drawback of the framework needs to be acknowledged, which is the representation of environment context. Using rasterized images divides the representations of agent state and scene context into two separate spaces, i.e., vector space (or discrete space) and image space. Moreover, the overly complex image space is unnecessary for driving environment representation and requires larger CNNs to process, as well as more computation resources and data to train the network. Therefore, future work will turn to the vectorized representation of environment \cite{gao2020vectornet}, which could provide a unified representation for the agent and environment, and significantly reduce the computation cost.

\section{Conclusions}
In this paper, we propose a deep learning framework named ReCoAt for multi-modal motion prediction, integrating recurrent neural networks for processing the target agent's historical motion states, convolutional neural networks for processing the environmental context in the format of rasterized images, as well as a novel attention mechanism for processing the interaction between agents. We train the framework on a real-world large-scale driving dataset covering a wide variety of urban scenarios and different kinds of target agents. We qualitatively demonstrate that ReCoAt is able to predict diverse, accurate, map-adaptive possible future trajectories for different target agents. In comparison against other methods on the standard testing benchmark, ReCoAt delivers the highest mean average precision, which measures both the prediction error and scoring error. In summary, ReCoAt achieves state-of-the-art performance, ranking the 2nd place winner of the 2021 Waymo Open Dataset Motion Prediction Challenge.

%%%%%%%%%%%%%%%%%%%%%%%%%%%%%%%%%%%%%%%%%%%%%%%%%%%%%%%%%%%%%%%%%%%%%%%%%%%%%%%%
\bibliographystyle{IEEEtran}
\bibliography{egbib}

\end{document}